\documentclass[letterpaper, 10 pt, conference]{ieeeconf}

\IEEEoverridecommandlockouts
\usepackage{graphics} 
\usepackage{epsfig}   
\usepackage{graphicx} 
\usepackage{iftex}
\ifPDFTeX
	\usepackage{mathptmx} 
	\usepackage{times}
\else
	\usepackage{fontspec}
\fi
\usepackage{amsmath}  
\usepackage{amssymb}  
\usepackage{cite}

\usepackage{booktabs}
\usepackage{multirow}
\usepackage{array}
\usepackage{tabularx}
\usepackage{caption}
\makeatletter
\g@addto@macro\normalsize{%
  \abovedisplayskip 3pt plus 2pt minus 1pt%
  \belowdisplayskip \abovedisplayskip%
  \abovedisplayshortskip 0pt plus 2pt%
  \belowdisplayshortskip 3pt plus 2pt minus 1pt%
}
\makeatother
\usepackage{xcolor}
\usepackage{float}
\usepackage{placeins}
\usepackage[hidelinks]{hyperref}

\newcolumntype{L}{>{\raggedright\arraybackslash}X}

\newif\ifcameraready
\camerareadytrue

\IEEEaftertitletext{\vspace{-6pt}}

\title{\LARGE \bf
SEA-Nav: Efficient Policy Learning for Safe and Agile Quadruped Navigation in Cluttered Environments 
}

\ifcameraready
\author{\vspace{-0.5em}
Shiyi Chen$^{1*}$, Mingye Yang$^{2*}$, Haiyi Liu$^{1*}$, Haiyan Mao$^{1}$, Jiaqi Zhang$^{1}$\\[0.25em]
Shuheng He$^{1}$, Debing Zhang$^{1\dagger}$, Zihao Qiu$^{1}$, Chun Zhang$^{1\dagger}$\\
Project Website: \href{https://11chens.github.io/sea-nav/}{\texttt{https://11chens.github.io/sea-nav/}}  
\thanks{*Equal contribution.}%
\thanks{$^{1}$Tsinghua University, Beijing, China.}%
\thanks{$^{2}$Imperial College London, London, UK.}%
\thanks{$^{\dagger}$Corresponding authors. Email: zhangchun@tsinghua.edu.cn}%
}
\else
\author{Anonymous Authors\\
Project Website: \href{https://sea-nav.github.io/}{\texttt{https://sea-nav.github.io/}}}
\fi

\begin{document}

\maketitle
\thispagestyle{empty}
\pagestyle{empty}

\begin{abstract}
Efficiently learning safe and agile quadruped navigation in densely cluttered environments remains difficult: existing methods often lack safety and agility, or become conservative in complex scenes and require long training schedules. We propose SEA-Nav (\underline{\textbf{S}}afe, \underline{\textbf{E}}fficient, and \underline{\textbf{A}}gile \underline{\textbf{Nav}}igation), a safe reinforcement learning framework for quadruped navigation in cluttered environments. A differentiable control barrier function (CBF) shield constrains the policy to produce safe velocity commands. An adaptive collision-state initialization mechanism increases the probability of learning from safety-critical near-collision experience. An action regularization term further suppresses infeasible commands for physical deployment. The policy converges after about one hour of training on a single RTX 4090 and transfers zero-shot to real-world cluttered scenes.
\end{abstract}

\section{INTRODUCTION}
Autonomous navigation in dense, cluttered environments remains a core challenge in robotics. Learning-based methods, especially imitation/self-supervised and imperative-learning paradigms, have shown strong local performance by fitting large-scale experience or learned cost priors \cite{gnm, badgr, ip, iplanner, viplanner, poliformer}. However, these methods often require expensive labeled data and can fail catastrophically under out-of-distribution (OOD) dense or dynamic obstacles.

To overcome static-dataset limitations, deep reinforcement learning (DRL) has been widely adopted for its maneuverability and reactivity \cite{abs, anymal_parkour, ETH_StateR, zhuang2023robot, cheng2023parkour, navrl, omniperception, reasan, BAS}. Yet RL in dense obstacles is difficult: obstacle penalties are hard to tune (large penalties cause over-conservative behavior, small penalties increase collisions), and low sample efficiency in long-horizon scenes makes it hard to balance goal reaching and aggressive avoidance. As a result, many methods need long training horizons or separate training stages \cite{abs, reasan, BAS}, increasing development and validation cost.

To inject explicit safety boundaries into RL, recent work integrates classical safety control methods such as Velocity Obstacles (VO) and Control Barrier Functions (CBF) \cite{DRL-VO, mult_VO_RL, CBFs, acbf_rl_rhc, comp_cbf, mult_const_cbf_sac, e_cbf}. However, existing schemes still compromise either end-to-end credit assignment (when used as post-processing filters) \cite{One_Filter, rl_safe1, rl_safe2} or stability under multiple constraints, where oscillation and conservative ``freezing'' can occur \cite{liao2023walking, One31, One32, One34}. More fundamentally, they do not solve the key bottleneck of high-quality experience sampling in dense-risk regions.

To address these issues, we propose an end-to-end safe RL framework for quadruped navigation in densely cluttered environments, combining efficient experience sampling with a differentiable CBF layer. The framework has three core components:

First, we introduce Adaptive Collision-State Initialization (ACSI) to improve sample utilization in high-density scenes. After a collision, the system probabilistically resets the robot to a critical pre-collision state in the local high-risk area. Combined with a goal-reaching-based reset curriculum, this focuses training on bottleneck regions and rapidly accumulates valuable avoidance experience.

\begin{figure}
\centering
\includegraphics[width=0.9\columnwidth]{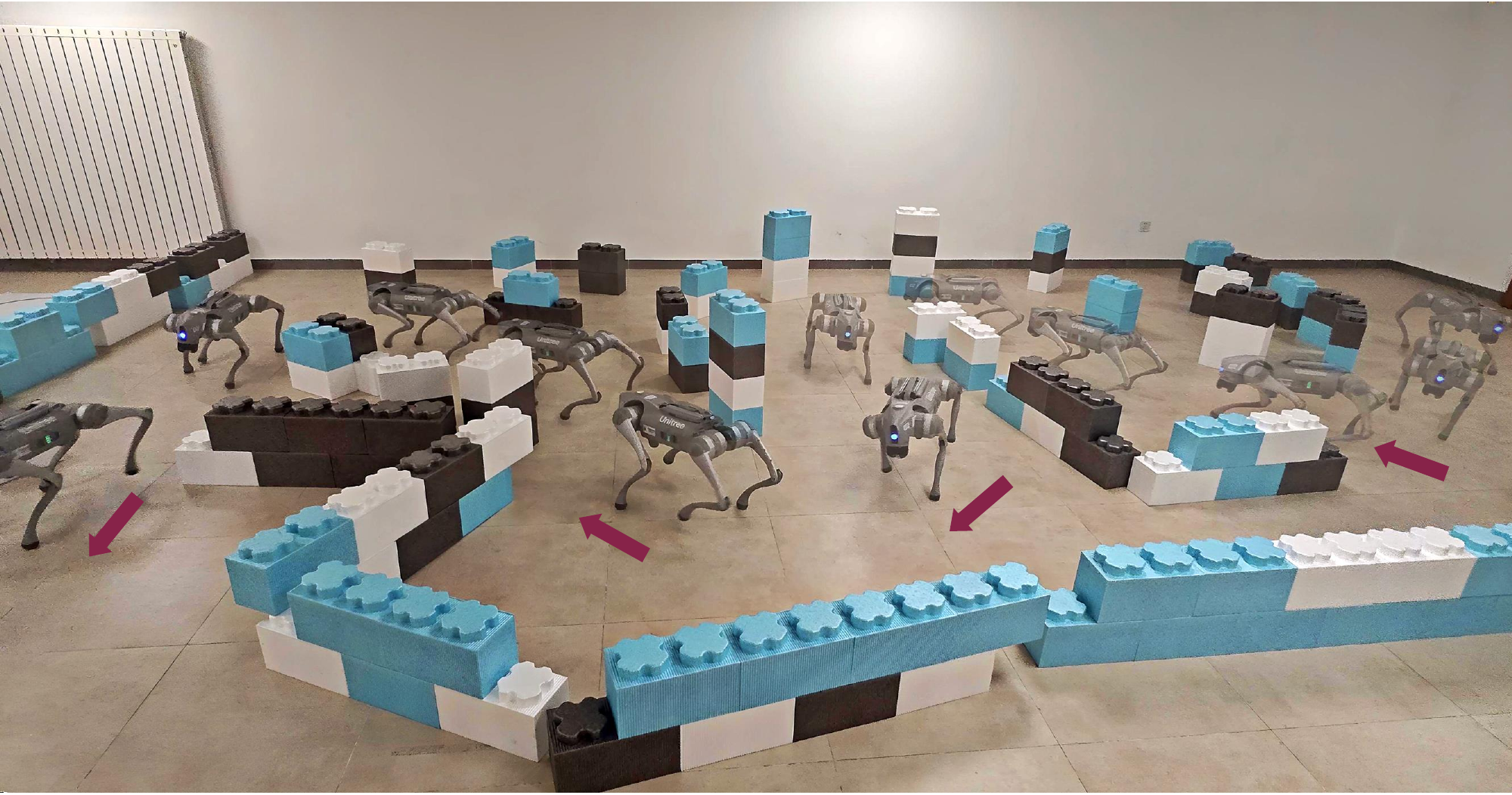}
\caption{SEA-Nav is trained in about one hour on a single RTX 4090 and deployed zero-shot in a previously unseen maze. The robot successfully escapes using the built-in MPC controller and onboard sparse LiDAR, both unseen during training.}
\label{fig:teaser}
\end{figure}

Second, we propose an end-to-end adaptive LSE-CBF safety projection layer. Instead of a rigid post-hoc filter that the policy never perceives during training, the shield is embedded in the training loop and provides an analytical geometric inductive bias in the action space, eliminating the train--test mismatch of post-processing safety schemes. We fuse multiple LiDAR constraints with Log-Sum-Exp (LSE) and add a physical damping term $\epsilon_d$ to avoid numerical singularities and the oscillatory control caused by discontinuous switching of active constraints in narrow passages. The closed-form projection remains differentiable, allowing the adaptive gain $\alpha$ to be optimized jointly with the navigation policy through the barrier layer.

Third, we introduce a kinematic action regularization loss that keeps velocity commands within physically executable bounds, facilitating direct deployment on real robots. With high sampling efficiency and differentiable safety constraints, our method achieves zero-shot, safe, and agile deployment in densely cluttered environments (Fig.~\ref{fig:teaser}).

The main contributions of this paper are summarized as follows:
\begin{itemize}
    \item \textbf{Adaptive Collision-State Initialization (ACSI):} A curriculum-guided critical-state replay strategy that addresses the sample-efficiency bottleneck of RL-based navigation in densely cluttered environments.
    \item \textbf{End-to-end Adaptive LSE-CBF Layer:} A closed-form differentiable safety layer trained in the loop with the policy, eliminating the train--test mismatch of post-hoc safety filters; LSE fusion with physical damping suppresses multi-constraint oscillation and adapts the avoidance aggressiveness online.
    \item \textbf{Deployment and Extensive Real-World Validation:} A kinematic action regularization loss that keeps learned commands physically executable, together with a deployment pipeline validated zero-shot on a Unitree Go2 across four real-world scenarios under two sensing--control configurations.
\end{itemize}

\section{RELATED WORK}

\subsection{Learning-Based Robot Navigation}
Early studies primarily relied on imitation learning and self-supervised learning (e.g., behavior cloning\cite{expert2, expert3, lv2023autonomous, shah2023vint}, environmental cost prediction\cite{amco, local_semantic, trav_costmap}) for end-to-end control or to assist traditional planners. Although these methods perform well in specific static scenarios, they are prone to collision failures when encountering OOD dense or dynamic obstacles. DRL has been widely applied to navigation tasks, demonstrating high maneuverability. However, pure RL faces two severe challenges in dense obstacles. First, sample efficiency is low \cite{poliformer, iplanner, PoliFormer_5}: since a collision immediately terminates the episode, the policy rarely dwells in the neighborhood of near-collision states, and safety-critical (near-collision) experience therefore constitutes only a negligible fraction of the collected samples. Second, pure RL lacks physical safety constraints, often leading to conservative behaviors under high collision penalties \cite{BAS}. These issues often translate into long training schedules before deployment across diverse environments. In RL-based quadruped navigation for cluttered environments specifically, representative systems illustrate this cost: ABS \cite{abs} couples an agile policy with a recovery controller switched by a separately learned reach-avoid value network, while REASAN \cite{reasan} and BAS \cite{BAS} likewise depend on multi-stage or multi-module training pipelines. Such designs increase training and validation overhead, and approximation errors may accumulate across module interfaces. Alternatively, OCR \cite{One_Filter} is a safety-filter approach that learns a single filter to post-process the commands of arbitrary downstream policies; since the policy never perceives the filter during training, a train--test mismatch arises between learned and deployed behavior. NavRL \cite{navrl} similarly enforces safety through a velocity-obstacle-based shield applied at execution time. Designing an efficient experience sampling mechanism while constraining physical safety from the action space remains a key challenge.

Recent reactive local-avoidance methods improve short-horizon responsiveness in cluttered scenes, but they still require careful integration with policy learning to maintain global efficiency and safety consistency \cite{reijgwart2024waverider}.

\subsection{Safe RL and Barrier Functions}
To provide explicit safety guarantees, prior works introduced VO as a post-processing safety shield. While VO can enforce hard kinematic constraints via optimization, it acts as a non-differentiable external filter that truncates gradient backpropagation. Moreover, its velocity-cone formulation causes severe conservatism in dense obstacle scenarios \cite{navrl}. Other studies attempted ``soft guidance'' by incorporating VO into the reward function \cite{DRL-VO, mult_VO_RL}, but this compromises strict analytical geometric constraints in the action space. 

In contrast, CBF formulations map spatial boundaries directly into control inputs via system dynamics, providing mathematically rigorous forward invariant set guarantees \cite{CBFs, acbf_rl_rhc, e_cbf}. More importantly, CBF formulations offer the potential for closed-form analytical solutions, paving the way for fully differentiable safety layers. Despite this potential, traditional multi-constraint CBFs aggregate constraints with the $\min$ operator; when the active constraint switches, the resulting control exhibits oscillations caused by the discontinuous change of the enforced boundary \cite{mult_const_cbf_sac}. Furthermore, static Class-$\mathcal{K}$ parameters ($\alpha$) fail to adapt to varying spatial densities, causing the robot to stall in narrow passages---an instance of the classical freezing-robot problem \cite{trautman2010unfreezing, liao2023walking}. Therefore, integrating adaptive parameters with smooth fusion mechanisms into a fully differentiable, optimizer-free analytical layer remains a critical challenge.
\section{METHOD}

\begin{figure*}[t]
    \centering
    \includegraphics[width=0.62\textwidth]{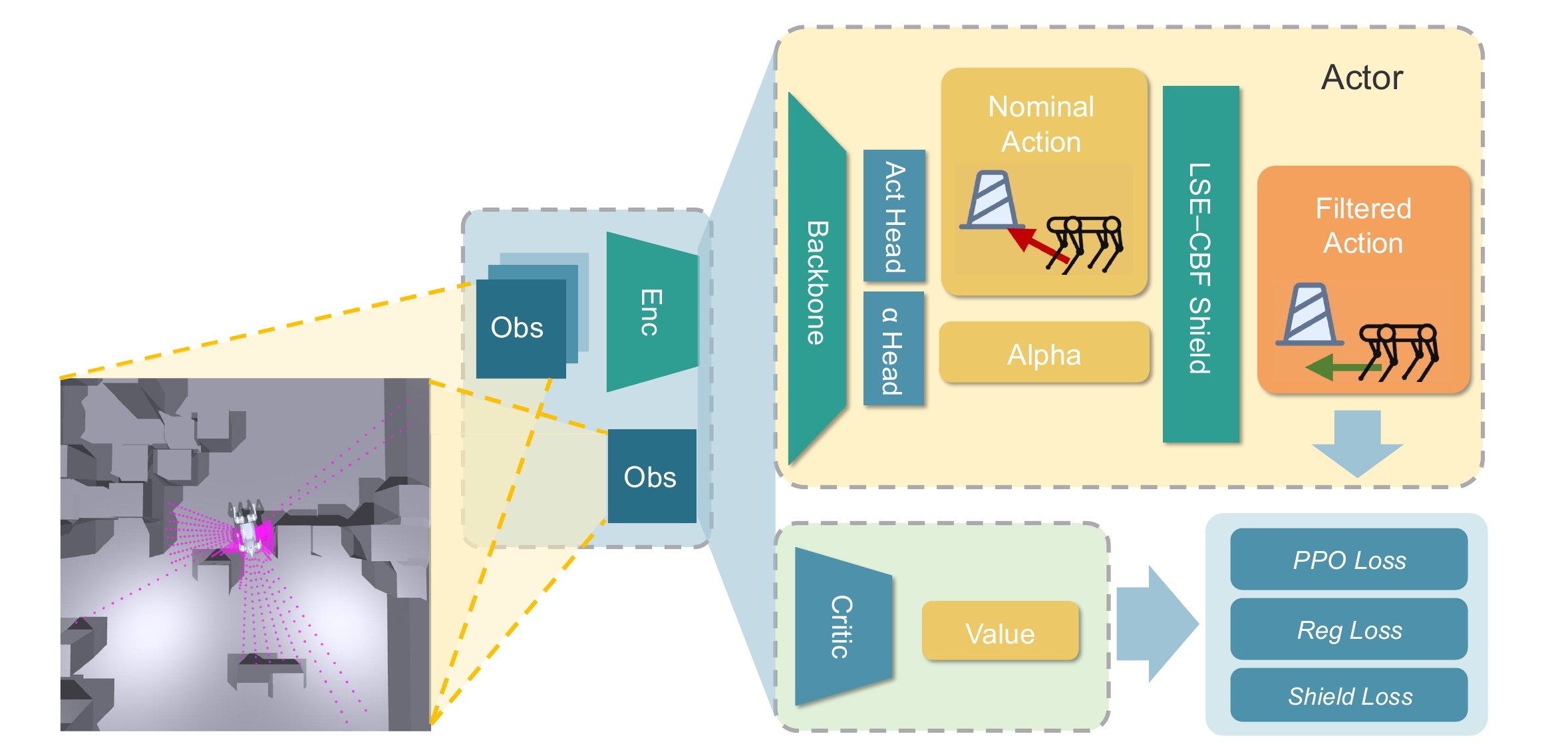}
    \vspace{0pt}
    \caption{Overview of the proposed SEA-Nav pipeline. Historical observations are encoded into a latent feature, which is concatenated with the current observation to form the joint state representation. In the Actor, the joint state passes through a shared Backbone into a navigation action head and a safety-gain $\alpha$ head, producing a nominal velocity command and an adaptive gain; the LSE-CBF Shield then solves for a safe velocity command. The Critic takes the joint state representation $x_t = [o_t, z_t]$ as input and directly predicts the state value $V(x_t)$. The Actor and Critic are optimized jointly with PPO, shield intervention, and kinematic regularization losses.}
    \label{fig:pipeline}
\end{figure*}

\subsection{Key Challenges \& Overview}
\textbf{Exploration-Exploitation Challenge in Dense Environments:} In densely cluttered environments, commonly used collision and reaching rewards are hard to balance: excessive collision penalties cause conservative movement, and immediate termination upon collision or goal reaching sharply reduces the proportion of critical experiences in narrow, high-risk regions and long-horizon tasks. Thus, we design ACSI to repeatedly replay critical trajectories before collisions, and set a goal-stay mechanism to deliberately prolong the learning of sparse experiences.

\textbf{Safe RL Constraints:} Without physical priors, pure RL training is inefficient and unsafe, while deploying safety barriers exclusively at test time introduces a severe train--test mismatch. A trainable barrier instead shapes the navigation policy during training, helping it learn safety awareness faster. Our solution relies on a Differentiable Barrier Layer based on Adaptive Multi-Constraint CBF.

\textbf{Robot Kinematic Constraints:} High-speed sharp turns are unsafe and can cause falls during deployment, while common action smoothing imposes hard constraints that severely restrict the action space. Thus, we design an action loss that constrains unsafe commands while preserving exploration.

\subsection{System Pipeline \& MDP Formulation}
This paper proposes a single-stage reinforcement learning framework that couples the Proximal Policy Optimization (PPO) algorithm with a differentiable CBF layer, as shown in Fig.~\ref{fig:pipeline}.

Our Actor-Critic architecture is designed to be end-to-end differentiable, and all neural networks within the architecture (including the Encoder, Backbone, respective Heads, and Critic) are composed of Multi-Layer Perceptrons (MLPs). In the Actor network, a sequence of historical observations $O_{hist} = \{o_{t-H}, \dots, o_{t-1}\}$ (where the history length $H = 10$ frames) first passes through the Encoder to extract a low-dimensional latent feature vector $z_t$. This latent representation is then concatenated with the current observation $o_t$ to form the joint state representation $x_t = [o_t, z_t]$, which is fed into a shared Backbone. The shared feature vector $f_t$ extracted by the Backbone is subsequently split into two parallel network heads: the \textbf{Navigation Action Head} outputs the desired nominal velocity command $\bar{u}_t = [\bar{v}_x, \bar{v}_y, \bar{\omega}_z]^T$; the \textbf{Safety Gain Head} ($\alpha$ Head) dynamically outputs the barrier gain parameter $\alpha_t$ (ensuring $\alpha_t > 0$ via a Softplus activation function). The outputs of both heads ($\bar{u}_t$ and $\alpha_t$) are fed into the LSE-CBF barrier layer to analytically compute the final safe command $u_{s,t} = [v_{x,s}, v_{y,s}, \omega_{z,s}]^T$, which drives the low-level locomotion controller. The Critic network, on the other hand, bypasses the safety barrier; it takes the joint state representation $x_t$ and passes it through an MLP to directly estimate the state value $V(x_t)$.

The Markov Decision Process (MDP) for the system interacting with the environment is formulated as follows:
\begin{itemize}
    \item \textbf{Observation Space:} The current observation $o_t \in \mathbb{R}^d$ includes the robot's base linear velocity $v^B \in \mathbb{R}^3$, base angular velocity $\omega^B \in \mathbb{R}^3$, projected gravity vector $g^B \in \mathbb{R}^3$, 2D local goal position $p_{goal}^B \in \mathbb{R}^2$ relative to the base frame, and a 2D LiDAR range scan $\rho \in \mathbb{R}^{41}$. The rays $\rho_i$ cover an angular range of $[-\frac{2\pi}{3}, \frac{2\pi}{3}]$ rad with a resolution of $\frac{\pi}{30}$ rad, and a sensing range of 0.1 m to 3.0 m. To align with real-world hardware deployment, the update frequency of exteroception (LiDAR $\rho$ and goal point $p_{goal}^B$) is set to 10 Hz, while proprioception ($v^B, \omega^B, g^B$) is maintained at 50 Hz.
    \item \textbf{Action Space:} The high-level navigation policy outputs the safe body velocity command $u_{s,t} \in \mathbb{R}^3$, which is sent to a pre-trained low-level locomotion controller at 50 Hz to be converted into target joint torques $\tau \in \mathbb{R}^{12}$.
    \item \textbf{Reward Design:} Our reward formulation centers on three key components: a clearance reward $r_{clear}$ to encourage traversing cluttered regions, a velocity reward $r_{velo}$ for goal-directed progress, and a stuck penalty $r_{stuck}$ to facilitate escape from local minima. Due to space constraints, the comprehensive formulation is detailed in Appendix~\ref{sec:appendix-reward}.
\end{itemize}

\subsection{Adaptive Collision-State Initialization (ACSI)}
To break the exploration bottleneck in high-density obstacle environments, we propose an adaptive collision replay mechanism. In conventional training, immediate termination upon collision followed by a reset to the initial state requires the robot to revisit the preceding route before returning to the failure region, limiting repeated learning of difficult avoidance situations.

ACSI instead keeps, for each environment, a rolling state buffer of the most recent $N_{buf}$ control steps (about 2\,s at 50\,Hz) containing the root state and the joint positions and velocities. When a collision occurs at step $t_{col}$, the episode is terminated early with the curriculum probability $P_{reset}$ defined below, and a termination penalty is applied so that the policy cannot exploit repeated collisions. With replay probability $p_{replay}$, the environment is then restored to the buffered state $s_{t_{col}-\Delta}$, retaining the true pose and velocities; otherwise, or if fewer than $\Delta_{min}$ steps of history exist, a standard reset is performed. The rollback length $\Delta$ is sampled uniformly and clipped by the buffer window and the elapsed episode length, so the robot is in practice restored to a state about 2\,s before the collision. These restored states form the local high-risk area, in which the policy retries the avoidance decision near the previous failure, allowing challenging avoidance situations in highly constrained spaces to be trained repeatedly.

To balance the accumulation of obstacle avoidance experience and global goal reaching, we introduce a reset curriculum based on goal-reaching performance. The probability $P_{reset}$ of collision-triggered early termination increases with the curriculum counter $L_{goal}$:
\begin{equation}\label{eq:acsi_curriculum}
P_{reset} = P_{min} + (P_{max} - P_{min}) \cdot \operatorname{clip}\left(L_{goal}/1.5,\, 0,\, 1\right)
\end{equation}
Here, $P_{min}$ and $P_{max}$ are the minimum and maximum early-termination probabilities. The curriculum counter $L_{goal}$ is updated according to the final goal distance $d$ as $L_{goal} \leftarrow L_{goal} + \mathbf{1}[d<d_{up}] - \mathbf{1}[d>d_{down}]$, where $\mathbf{1}[\cdot]$ denotes the indicator function, and $d_{up}$ and $d_{down}$ are the distance thresholds for increasing and decreasing the counter, respectively.

This curriculum guides the robot to prioritize exploration and goal reaching early in training, while shifting its focus to avoidance in high-risk areas as goal-reaching performance improves: a low early-termination probability lets exploration continue after collisions, whereas a high probability provides more opportunities for pre-collision restoration and local avoidance learning. All ACSI hyperparameters are listed in Table~\ref{table:train_hyperparams} (Appendix~\ref{sec:appendix-hyperparams}).

\subsection{Differentiable Adaptive LSE-CBF Layer}

To enhance the safety of the policy network in complex environments, this module introduces a fully differentiable multi-constraint CBF layer. 

\subsubsection{LSE Aggregation for Smooth Safety Constraints}
In our LiDAR-based navigation task, the safety set $\mathcal{C}$ is defined by the intersection of constraints from $N=41$ discrete rays. A straightforward formulation uses $h(x) = \min_{i} h_i(x)$ with per-ray barriers $h_i(x) = \rho_i - d_{safe}$, where $\rho_i$ is the $i$-th ray range and $d_{safe}$ is the safety clearance (the robot safety radius plus a small margin). The non-differentiable $\min$ causes gradient jumps during constraint switching. For instance, in a narrow corridor, a marginal shift in proximity (e.g., from a $0.50$\,m left wall to a $0.51$\,m right wall) forces the active constraint's spatial gradient to abruptly flip $180^\circ$, triggering directional chattering. To address this, we adopt the Log-Sum-Exp (LSE) formulation, a standard smooth approximation in safe control \cite{mult_const_cbf_sac, comp_cbf}, to fuse all discrete constraints into a global composite safety function $h(x)$:
\begin{equation}\label{eq:lse_cbf}
    h(x) = -\frac{1}{k} \ln \left( \sum_{i=1}^{N} \exp(-k \cdot h_i(x)) \right)
\end{equation}
where $k$ is the smoothing coefficient; we implement \eqref{eq:lse_cbf} in a min-shifted, numerically stable form. This formulation ensures that $h(x)$ is continuously differentiable, providing a smooth gradient landscape for both control execution and neural network backpropagation.

\subsubsection{Damped Analytical Safety Projection}
Given the smoothed safety margin $h(x)$, the CBF layer modulates the Actor's nominal action $\bar{u}(x)$ to satisfy the forward invariance condition $\dot{h}(x, u) \ge -\alpha h(x)$. This requirement yields the standard CBF-QP \cite{Ames_2017, comp_cbf}:
\begin{equation}\label{eq:qp_cbf}
    \min_{u_s} \frac{1}{2} || u_s - \bar{u}(x) ||^2 \quad \text{s.t.} \quad \langle L_g h(x), u_s \rangle + \alpha h(x) \ge 0
\end{equation}
Here we model the base as a 2D single integrator (point mass): the high-level command acts directly as the planar velocity, so $L_f h = 0$ and $L_g h_i(x) = -n_i$ with $n_i$ the unit direction of the $i$-th ray; $L_g h(x) = \nabla h(x)$ then represents the gradient direction of the safety margin, and $\alpha > 0$ is the Class-$\mathcal{K}$ parameter. The projection corrects only the planar components $(v_x, v_y)$, while the yaw rate $\omega_z$ passes through unmodified. This approximation is adequate for high-level planning because the low-level locomotion controller is responsible for tracking the commanded velocity, so whole-body dynamics need not be modeled at this layer.

While LSE resolves directional chattering by smoothly blending gradients, it introduces a challenge in perfectly symmetric environments. In narrow passages where opposing hazard gradients (e.g., from left and right walls) equally blend and cancel each other out, the composite gradient vanishes ($||L_g h(x)||^2 \to 0$). The closed-form solution of \eqref{eq:qp_cbf} in \cite{comp_cbf} then yields a vanishing denominator, causing the correction magnitude to grow unboundedly and leading to numerical divergence. To resolve this, we introduce a physical damping term $\epsilon_d$ in the closed-form solution, resulting in the modified safety command $u_s(x)$:
\begin{equation}\label{eq:analytical_cbf}
\begin{aligned}
    u_s(x) =\; & \bar{u}(x) + \\
    &\underbrace{\max \left\{ 0, \frac{-(\langle L_g h(x), \bar{u}(x) \rangle + \alpha \cdot h(x))}{||L_g h(x)||^2 + \epsilon_d} \right\}}_{\text{Correction Magnitude } \eta} \cdot \underbrace{L_g h(x)}_{\text{Direction}}
\end{aligned}
\end{equation}
where the $\max\{0, \cdot\}$ operator acts as a logical switch: it leaves safe actions unaltered ($\eta=0$), while generating a strictly positive magnitude $\eta$ to project unsafe actions along the safety gradient $L_g h(x)$. Note the asymmetry with the $\min$ aggregation: under $\min$, a switch of the active constraint flips the gradient \emph{direction}, whereas $\max\{0,\cdot\}$ only scales a nonnegative \emph{magnitude}. The operator is piecewise linear and subdifferentiable and never alters the correction direction, so the problematic directional jump does not arise. The damping term $\epsilon_d$ caps the maximum correction magnitude, preventing unbounded velocity corrections when gradients vanish in highly constrained spaces. With $\epsilon_d>0$, \eqref{eq:analytical_cbf} is a damped approximation of the exact CBF-QP minimizer. The damping term trades strict constraint enforcement for numerical stability.

\subsubsection{End-to-End Differentiability as Inductive Bias}
Owing to the LSE smoothing, the $\epsilon_d$ damping, discrete-time execution, sensor noise, and low-level tracking errors, this layer does not provide a certified safety guarantee; it serves as a \textbf{differentiable safety inductive bias}. Since \eqref{eq:analytical_cbf} is composed entirely of differentiable operations, the adaptive gain $\alpha$ can be optimized jointly with the navigation policy through the barrier layer. This allows the network to autonomously learn to increase $\alpha$ (be aggressive) in open areas and decrease $\alpha$ (be conservative) in narrow spaces, internalizing physical safety constraints into the policy itself.

\begin{figure}[t]
    \centering
    \includegraphics[width=\columnwidth]{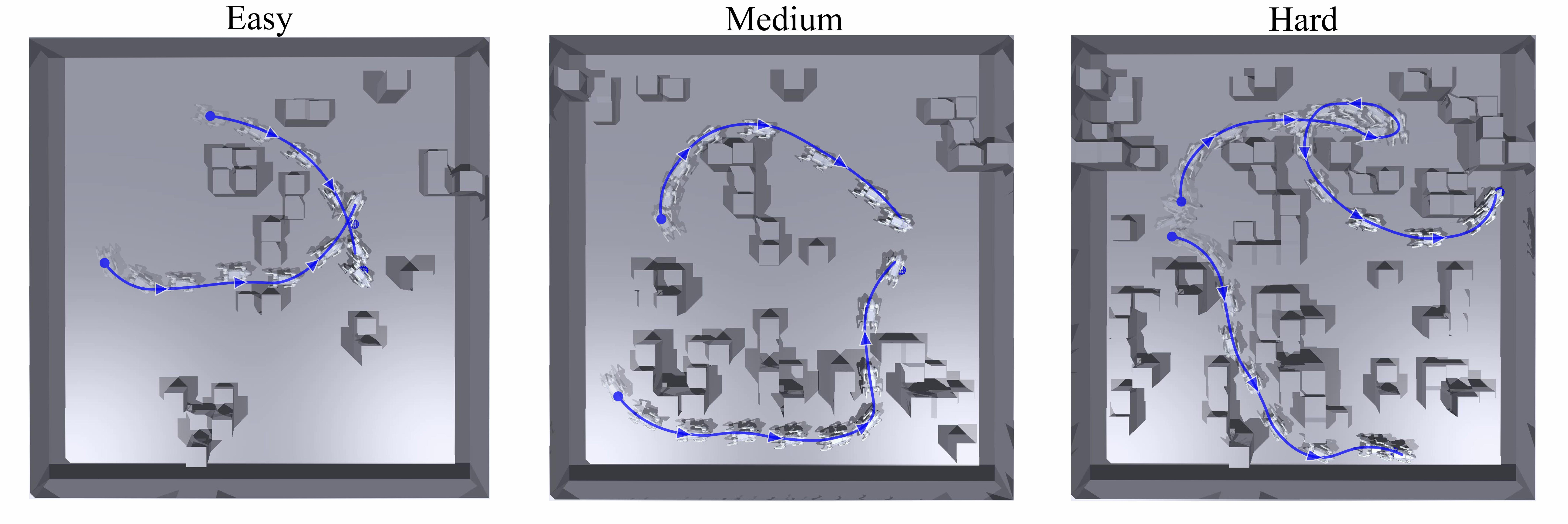}
    \caption{SEA-Nav trajectories in three navigation scenarios of increasing difficulty. Each subfigure shows two distinct start-goal trials. SEA-Nav successfully traverses narrow passages and performs timely maneuvering adjustments to avoid entrapment in cluttered environments.}
    \label{fig:total_rooms}
\end{figure}

\begin{table*}[htbp]
\centering
\scriptsize
\setlength{\tabcolsep}{2.6pt}
\renewcommand{\arraystretch}{0.95}
\caption{Different difficulty levels are evaluated; mean and standard deviation are reported over 3 evaluation runs of 100 randomized trials each.}
\label{tab:sim_results}
\begin{tabular*}{\textwidth}{@{\extracolsep{\fill}}>{\raggedright\arraybackslash}p{0.15\textwidth}ccc@{\hskip 15pt}ccc@{\hskip 15pt}ccc}

\toprule
\multirow{2}{*}{\textbf{Method}}
& \multicolumn{3}{c}{Easy} 
& \multicolumn{3}{c}{Medium} 
& \multicolumn{3}{c}{Hard} \\ 
\cmidrule(lr){2-4} \cmidrule(lr){5-7} \cmidrule(lr){8-10}
& SR $(\%)$ $\uparrow$ & CR $(\%)$ $\downarrow$ & TR $(\%)$ $\downarrow$ 
& SR $(\%)$ $\uparrow$ & CR $(\%)$ $\downarrow$ & TR $(\%)$  $\downarrow$ 
& SR $(\%)$ $\uparrow$ & CR $(\%)$ $\downarrow$ & TR $(\%)$ $\downarrow$ \\

\midrule
\textbf{SEA-Nav (Ours)}                
& \textbf{100.00} \textcolor{gray}{\tiny(±0.00)} & \textbf{0.00} \textcolor{gray}{\tiny(±0.00)} & \textbf{0.00} \textcolor{gray}{\tiny(±0.00)}
& \textbf{97.00} \textcolor{gray}{\tiny(±0.82)} & \textbf{1.00} \textcolor{gray}{\tiny(±0.82)} & \textbf{2.00} \textcolor{gray}{\tiny(±0.00)}  
& \textbf{90.00} \textcolor{gray}{\tiny(±1.63)} & \textbf{5.00} \textcolor{gray}{\tiny(±0.82)} & \textbf{5.00} \textcolor{gray}{\tiny(±0.82)} \\

\midrule
\textbf{SEA-Nav w/o ACSI}                
& \textbf{100.00} \textcolor{gray}{\tiny(±0.00)} & \textbf{0.00} \textcolor{gray}{\tiny(±0.00)} & \textbf{0.00} \textcolor{gray}{\tiny(±0.00)}
& 93.00 \textcolor{gray}{\tiny(±0.82)} & 2.33 \textcolor{gray}{\tiny(±0.47)} & 4.67 \textcolor{gray}{\tiny(±0.47)}  
& 83.00 \textcolor{gray}{\tiny(±2.45)} & 8.00 \textcolor{gray}{\tiny(±2.16)} & 9.00 \textcolor{gray}{\tiny(±1.63)} \\

\midrule
\textbf{SEA-Nav w/o Shield} 
& 97.67 \textcolor{gray}{\tiny(±0.47)} & 1.33 \textcolor{gray}{\tiny(±0.47)} & 1.00 \textcolor{gray}{\tiny(±0.00)}  
& 90.00 \textcolor{gray}{\tiny(±0.82)} & 2.67 \textcolor{gray}{\tiny(±0.94)} & 7.33 \textcolor{gray}{\tiny(±1.70)} 
& 74.33 \textcolor{gray}{\tiny(±3.30)} & 11.67 \textcolor{gray}{\tiny(±1.70)} & 14.00 \textcolor{gray}{\tiny(±2.16)} \\

\midrule
\textbf{SEA-Nav w/o $L_{reg}$}  
& 96.33 \textcolor{gray}{\tiny(±0.47)} & 2.67 \textcolor{gray}{\tiny(±0.47)} & 1.00 \textcolor{gray}{\tiny(±0.00)}  
& 82.33 \textcolor{gray}{\tiny(±2.05)} & 6.00 \textcolor{gray}{\tiny(±0.82)} & 11.67 \textcolor{gray}{\tiny(±1.25)} 
& 57.00 \textcolor{gray}{\tiny(±4.08)} & 18.00 \textcolor{gray}{\tiny(±2.16)} & 25.00 \textcolor{gray}{\tiny(±2.16)} \\

\midrule
\textbf{ABS\cite{abs}}                
& 95.00 \textcolor{gray}{\tiny(±0.00)} & 4.67 \textcolor{gray}{\tiny(±0.47)} & 0.33 \textcolor{gray}{\tiny(±0.47)}
& 75.33 \textcolor{gray}{\tiny(±1.25)} & 9.33 \textcolor{gray}{\tiny(±0.47)} & 15.33 \textcolor{gray}{\tiny(±1.70)}  
& 45.33 \textcolor{gray}{\tiny(±3.40)} & 22.00 \textcolor{gray}{\tiny(±1.41)} & 32.67 \textcolor{gray}{\tiny(±2.05)} \\

\midrule
\textbf{OCR\cite{One_Filter}}                
& 95.33 \textcolor{gray}{\tiny(±0.47)} & 3.33 \textcolor{gray}{\tiny(±0.47)} & 1.33 \textcolor{gray}{\tiny(±0.47)}
& 67.33 \textcolor{gray}{\tiny(±2.49)} & 31.33 \textcolor{gray}{\tiny(±2.87)} & 32.67 \textcolor{gray}{\tiny(±2.49)}  
& 56.00 \textcolor{gray}{\tiny(±3.27)} & 18.33 \textcolor{gray}{\tiny(±2.49)} & 25.67 \textcolor{gray}{\tiny(±2.49)} \\

\midrule
\textbf{REASAN\cite{reasan}}                
& 95.67 \textcolor{gray}{\tiny(±0.47)} & 2.67 \textcolor{gray}{\tiny(±0.47)} & 1.33 \textcolor{gray}{\tiny(±0.47)}
& 71.33 \textcolor{gray}{\tiny(±3.09)} & 10.33 \textcolor{gray}{\tiny(±1.89)} & 28.67 \textcolor{gray}{\tiny(±2.16)}  
& 77.67 \textcolor{gray}{\tiny(±4.62)} & 14.33 \textcolor{gray}{\tiny(±1.89)} & 22.33 \textcolor{gray}{\tiny(±4.19)} \\

\bottomrule
\end{tabular*}
\end{table*}

\subsection{Loss Function Design}
To foster a synergistic relationship between the policy network and the LSE-CBF layer, we introduce a \textbf{Shield Intervention Loss} ($L_{shield}$). This loss minimizes the discrepancy between the nominal command $\bar{u}_t$ and the shielded command $u_{s,t}$, while discouraging overly small $\alpha$ that would trigger near-complete shield intervention and introduce dynamics risks:
\begin{equation}
L_{shield} = ||u_{s,t} - \bar{u}_t||^2 + [\alpha_{min} - \alpha_t]_+^2
\end{equation}
where $[x]_+ = \max\{0, x\}$ and $\alpha_{min}$ is a lower bound for the adaptive gain.

Furthermore, since the execution performance of quadruped locomotion is highly sensitive to the navigation commands, we aggregate physical constraints and network smoothness into a unified \textbf{Kinematic Regularization Loss} ($L_{reg}$). This regularization is composed of two sub-components: a range penalty ($L_{range}$) and a smoothness penalty ($L_{smooth}$). 

First, to prevent destructive behaviors, the velocity command range loss penalizes outputs that exceed the hardware safety limits:
\begin{equation}
    L_{range} = \sum_{j \in \{x,y,\omega\}} \left( u_{s,t}^j - \operatorname{clip}(u_{s,t}^j, u^{\min}_j, u^{\max}_j) \right)^2
\end{equation}
where $u^{\min}, u^{\max} \in \mathbb{R}^{3}$ are the element-wise lower and upper bounds of the robot's kinematic capabilities. 

Second, to promote smooth transitions in both action and value predictions, we adopt a Lipschitz continuity constraint \cite{L2C2}:
\begin{equation}
L_{smooth} = \lambda_\pi D\left(\pi_\theta(x_t), \pi_\theta(\bar{x}_t)\right) + \lambda_V D\left(V_\phi(x_t), V_\phi(\bar{x}_t)\right)
\end{equation}
where $\bar{x}_t = x_t + \beta \cdot (x_{t+1} - x_t)$ is an interpolated state with the coefficient $\beta \sim \mathcal{U}(-1,1)$, and $D(\cdot, \cdot)$ denotes the mean squared error (MSE). Here, $\pi_\theta$ and $V_\phi$ represent the policy and value networks taking the joint state representation $x_t$ as input, while $\lambda_\pi$ and $\lambda_V$ are their corresponding weighting factors. This constraint improves deployment safety by suppressing abrupt changes in the action outputs, reducing the risk of falls and motor overheating during Sim-to-Real locomotion. In addition, the value term regularizes the critic and stabilizes value learning, which explains why removing $L_{reg}$ degrades not only collision avoidance but also overall task completion (Table~\ref{tab:sim_results}). The total kinematic regularization is thus given by $L_{reg} = L_{range} + L_{smooth}$.

Finally, these auxiliary losses are integrated with the standard PPO objective \cite{ppo}. The total loss function used to update the network is formulated as:
\begin{equation}
    L_{total} = L_{PPO} + \lambda_{shield} L_{shield} + \lambda_{reg} L_{reg}
\end{equation}
where $\lambda_{shield}$ and $\lambda_{reg}$ are the weighting hyperparameters for each respective loss term.

\section{EXPERIMENTS}

\begin{figure*}[htbp]
    \centering
    \includegraphics[width=0.72\textwidth]{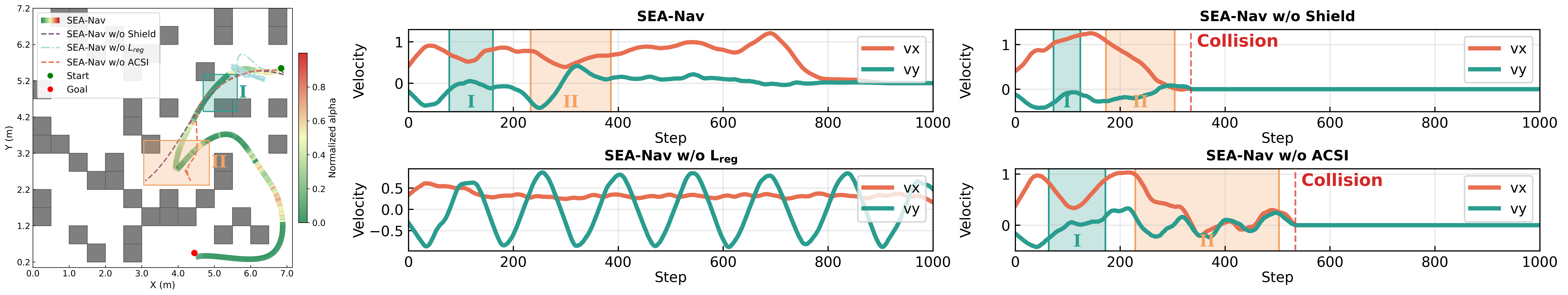}
    \caption{(Left) Trajectory plot; (right) velocity profiles. SEA-Nav maintains larger obstacle clearance and smoother speed variations. The safety gain $\alpha$ decreases in hazardous regions so the CBF Shield dominates, and increases in safe regions where the nominal navigation command takes the lead.}
    \label{fig:case_vel}
\end{figure*}

\begin{figure}[t]
    \centering
    \includegraphics[width=0.80\columnwidth]{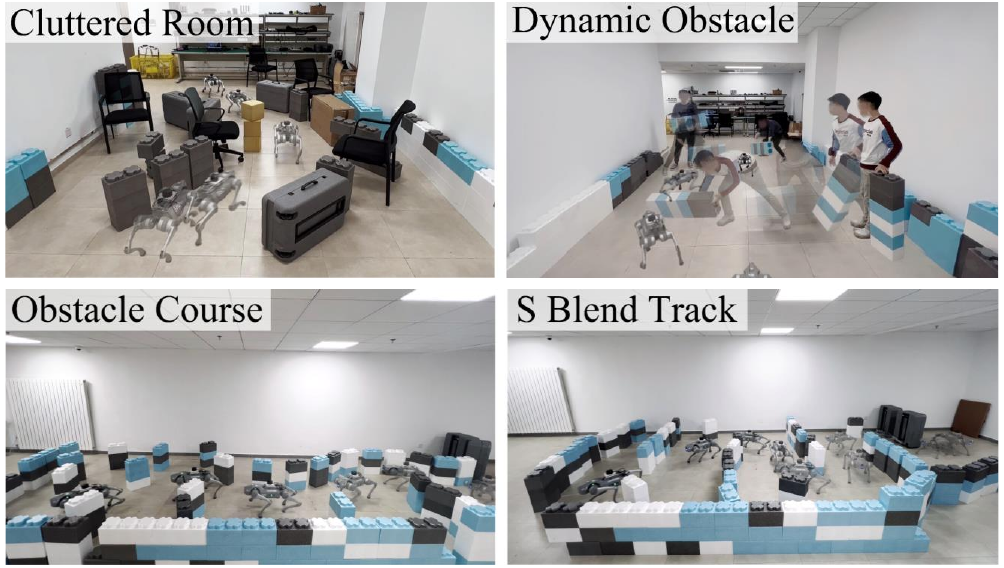}
    \caption{Real-world experimental environments. Ten trials are conducted in each environment.}
    \label{fig:real_env}
\end{figure}

\begin{table*}[htbp]
\centering
\scriptsize
\setlength{\tabcolsep}{3.2pt}
\renewcommand{\arraystretch}{0.95}
\caption{Real-world deployment results. AS: average speed (m/s); SEA-Nav-b: built-in MPC controller and onboard sparse LiDAR.}
\label{table:realworld-results}
\resizebox{\textwidth}{!}{%
\begin{tabular}{lcccccccccccc}
\toprule
\multirow{2}{*}{\textbf{Method}}
& \multicolumn{3}{c}{Cluttered Room}
& \multicolumn{3}{c}{Dynamic Obstacle}
& \multicolumn{3}{c}{Obstacle Course}
& \multicolumn{3}{c}{S-Blend Track} \\
\cmidrule(lr){2-4} \cmidrule(lr){5-7} \cmidrule(lr){8-10} \cmidrule(lr){11-13}
& SR $(\%)$ $\uparrow$ & CR $(\%)$ $\downarrow$ & AS (m/s) $\uparrow$
& SR $(\%)$ $\uparrow$ & CR $(\%)$ $\downarrow$ & AS (m/s) $\uparrow$
& SR $(\%)$ $\uparrow$ & CR $(\%)$ $\downarrow$ & AS (m/s) $\uparrow$
& SR $(\%)$ $\uparrow$ & CR $(\%)$ $\downarrow$ & AS (m/s) $\uparrow$ \\
\midrule
SEA-Nav              & \textbf{100} & 10 & 1.6 & 90 & 20 & 1.5 & \textbf{100} & 10 & \textbf{1.2} & \textbf{100} & 10 & \textbf{1.4} \\
SEA-Nav-b            & \textbf{100} & 10 & 0.9 & \textbf{100} & 20 & 0.8 & 90 & 10 & 0.7 & \textbf{100} & 10 & 0.5 \\
ABS\cite{abs}        & 70 & 50 & \textbf{2.1} & 60 & 10 & \textbf{1.7} & 0 & 100 & 0.6 & 80 & 20 & 0.8 \\
OCR\cite{One_Filter} & 90 & 20 & 1.7 & 90 & 10 & 1.5 & 30 & 100 & 1.0 & 90 & 20 & 1.3 \\
REASAN\cite{reasan}  & 90 & 20 & 1.3 & 90 & 10 & 1.3 & 40 & 100 & 0.9 & 90 & 30 & 0.9 \\
SLAM                 & \textbf{100} & \textbf{0} & 0.5 & 80 & 40 & 0.4 & \textbf{100} & \textbf{0} & 0.4 & 80 & \textbf{0} & 0.3 \\
\bottomrule
\end{tabular}%
}
\end{table*}

\subsection{Simulation and Training Setup}
Using the Isaac Gym platform \cite{lggym}, we first trained a standard velocity-tracking controller based on \cite{slr}. For the navigation task, we designed $10\,\text{m} \times 10\,\text{m}$ rooms heavily occupied by diverse obstacles (Fig.~\ref{fig:total_rooms}), exposing the policy to varied obstacle layouts during training. An episode ends only after the robot stays near the goal for a period, increasing the proportion of goal-reaching experiences, and the collision bodies are slightly expanded to enhance collision perception and leave an additional safety margin for real-world deployment. All policies are trained with 2048 parallel environments and 48 steps per iteration for 2000 PPO iterations ($\approx 1.97\times10^{8}$ environment steps), converging in about one hour on a single RTX 4090 GPU. All ablation variants are trained with the identical configuration and this fixed budget, and all results are reported at convergence. The w/o Shield variant removes the entire differentiable CBF layer together with the adaptive-gain head rather than replacing the LSE aggregation with a hard $\min$. All baselines are trained from their official implementations under our training configuration, with the same environments and budget.

\subsection{Simulation Results}
Evaluation rooms are procedurally generated from random seeds at three occupancy difficulties (Easy, Medium, and Hard; Fig.~\ref{fig:total_rooms}), so each difficulty corresponds to a family of layouts rather than a fixed handcrafted scene. We conducted three independent evaluation runs of 100 trials per difficulty (300 trials per difficulty), with randomized start points, end points, and initial headings. We report Success Rate (SR), Collision Rate (CR), and Timeout Rate (TR), using the same definitions for all methods. CR measures the proportion of trials with at least one collision, including trials that subsequently succeed or time out. TR measures the proportion of trials that exceed the 30\,s evaluation limit without reaching the goal.

Our training design achieves a favorable balance among agility, safety, and task completion, especially in hard environments. The ablation results in Table \ref{tab:sim_results} show that ACSI improves success by repeatedly revisiting high-risk states, which increases near-obstacle experience and reduces collision-induced failures. Since collision-triggered resets keep near-collision samples inherently rare, ACSI changes the training data distribution rather than the training speed; simply training the w/o ACSI variant longer would mainly accumulate low-value samples in open regions without resolving this scarcity. Removing the differentiable Shield degrades both safety and task completion, indicating that shaping the action space with the LSE-CBF layer provides more efficient avoidance learning and a stronger safety bias. The kinematic regularization further guides the policy toward hardware-feasible velocity commands, improving stability and deployment robustness.

\subsection{Case Study}
We further analyze the contributions of ACSI, the Shield, and the kinematic regularization in a randomly generated scene unseen during training: a long-horizon scenario requiring successive avoidance maneuvers and containing local minima. Fig.~\ref{fig:case_vel} visualizes the trajectories and velocity profiles. With the full method, the adaptive gain $\alpha$ decreases when the robot enters narrow passages, increasing shield intervention and enabling safe passage; the turning phase also exhibits smooth, bounded velocity transitions. The stuck-aware reward helps the robot escape local minima, while the regularization loss suppresses aggressive velocity spikes that would otherwise lead to unsafe impacts. In contrast, removing the regularization yields high-frequency velocity oscillations (most visible in $v_y$), which on hardware would increase the risk of falls and motor overheating; removing ACSI or the Shield reduces exposure to high-risk states or removes explicit safety shaping, consistent with the higher collision rates in Table~\ref{tab:sim_results}. Note that $L_{reg}$ is not equivalent to simply lowering the command limit: $L_{range}$ penalizes only violations of the hardware bounds themselves, and $L_{smooth}$ constrains the sensitivity of the outputs over time rather than their magnitude. Suppressing transient spikes by reducing $v_{max}$ would sacrifice agility, whereas SEA-Nav sustains average speeds of 1.2--1.6\,m/s (Table~\ref{table:realworld-results}).

\subsection{Real-World Setup}
We evaluated our method on the Unitree Go2 quadruped robot across different obstacle environments. Among the four scenarios (Fig.~\ref{fig:real_env}), the Cluttered Room and the Dynamic Obstacle arena measure approximately $2.4\,\text{m} \times 4.8\,\text{m}$, while the Obstacle Course and the S-Blend Track measure approximately $4.2\,\text{m} \times 10.2\,\text{m}$. The narrowest passage is 0.6\,m wide, roughly twice the 0.31\,m body width of the Go2, leaving little lateral clearance. In the Dynamic Obstacle scenario, a pedestrian crosses the robot's path laterally at about 1.5\,m/s, starting when the robot is roughly 1\,m away. Each method is evaluated for 10 trials per scenario with randomized initializations (Table~\ref{table:realworld-results}). We designed two deployment schemes for SEA-Nav:
\begin{enumerate}
    \item Using the robot's onboard sparse LiDAR L1 for perception and the built-in MPC controller (max 1 m/s).
    \item Using a higher-precision RPLIDAR A2 for perception and our trained agile policy \cite{slr} as the controller.
\end{enumerate}
Different from prior LiDAR-based approaches \cite{One_Filter, omniperception, reasan}, which require high-precision ray measurements and do not support the stock Unitree L1 LiDAR, our first scheme uses the native sensor and controller to enable a plug-and-play, low-cost deployment path. We summarize the point-cloud differences between the two LiDAR setups in Appendix~\ref{sec:appendix-pointcloud} (Fig.~\ref{fig:pointcloud}). We also compare with the built-in SLAM navigation of the Unitree Go2.

\subsection{Real-World Results}

Table~\ref{table:realworld-results} summarizes the quantitative results. Our method achieves safer obstacle avoidance than state-of-the-art baselines in the most difficult navigation environments. It remains reliable in continuous turning and local planning. In contrast, while some baselines move faster, they either become stuck or crash at dangerous speeds when facing narrow corners. The SLAM method has a high success rate but relies on lengthy mapping, moves slowly, and struggles with dynamic obstacles (CR 40\%). The zero-shot transfer is supported by three factors: domain randomization over dynamics and observations during training (Appendix~\ref{sec:appendix-domain-random}), an exteroceptive update rate (10\,Hz) aligned with the real LiDAR, and the fact that the same policy operates under two distinct perception--control combinations (sparse L1 with the built-in MPC, and RPLIDAR A2 with a learned controller), indicating robustness to both perception and execution discrepancies.

\section{CONCLUSION}
In this paper, we proposed SEA-Nav, an efficient RL navigation framework for quadruped robots that integrates ACSI experience replay and an end-to-end differentiable LSE-CBF shield. It achieves agile and safe deployment with about one hour of training.

\textbf{Limitations:} The current algorithm only supports flat-ground navigation and lacks detection capabilities for slopes or stairs. It can overcome simple local optima but is still prone to getting stuck in complex mazes and dead ends. 

\textbf{Future Work:} We plan to incorporate global navigation algorithms or memory mechanisms to resolve complex local optima and expand terrain adaptability.

\appendix
\subsection{Reward Function}
\label{sec:appendix-reward}

Table~\ref{tab:reward-terms} summarizes the reward terms and weights used in our implementation. Let $d$ denote the goal distance; $\mathbf{v}=[v_x,v_y]$ the base linear velocity in the horizontal plane; and $\boldsymbol\omega=[\omega_x,\omega_y,\omega_z]$ the base angular velocity. $\mathbf{1}[\cdot]$ is the indicator function. $\theta$ is the heading error to the goal. $\phi$ is the angular offset to the most open ray direction, and $c_{\text{front}}$ is the minimum ray clearance in a frontal cone. $\rho_i$ is the $i$-th ray distance. For collisions, $\mathbf{f}_k$ is the contact force magnitude on body group $k$, and $\beta_k$ are body-dependent coefficients (omitted for brevity). For stuck detection, we track recent planar base positions $\{\mathbf{p}_t\}_{t=1}^T$ and define $\Delta p_{\max}=\max_t\lVert\mathbf{p}_t-\mathbf{p}_1\rVert_2$.

\begin{table}[htbp]
\centering
\footnotesize
\setlength{\tabcolsep}{5pt}
\renewcommand{\arraystretch}{1.0}
\caption{Reward terms and weights.}
\label{tab:reward-terms}
\resizebox{\columnwidth}{!}{%
\begin{tabular}{l l l r}
\toprule
\textbf{Term} & \textbf{Purpose} & \textbf{Compact expression} & \textbf{Weight} \\
\midrule
$ r_{\text{term}} $ & episode termination cost & $\mathbf{1}[\text{terminated}]$ & $-100$ \\

$ r_{\text{reach}} $ & tight goal reaching & $\frac{1}{1+2d^2}\,\mathbf{1}[d<0.5]$ & $+10$ \\

$ r_{\text{velo}} $ & move toward goal & $\cos\theta\,v_x+\frac{1}{1+2d^2}$ & $+15$ \\

$ r_{\text{clear}} $ & clearance-aware motion & $\mathbf{1}[d>1] \,\cos\phi\,v_x+\mathbf{1}[d\le 1] \,\frac{1}{1+2d^2}$ & $+15$ \\

$ r_{\text{stuck}} $ & penalize getting stuck & $\mathbf{1}[d>1] \,\mathbf{1}[\Delta p_{\max}<0.1] \,\mathbf{1}[v_x>0.0] \,\mathbf{1}[|\omega_z|<1.0]$ & $-5$ \\

$ r_{\text{coll}} $ & penalize collisions & $\big(1+4(\lVert\mathbf{v}\rVert_2^2+\omega_z^2)\big)\,\sum_{k}\beta_k\,\mathbf{1}[\lVert\mathbf{f}_k\rVert>0.1]$ & $-4$ \\

$ r_{\omega} $ & reduce angular motion & $\lVert[\omega_x,\omega_y]\rVert_2$ & $-0.05$ \\
\bottomrule
\end{tabular}%
}
\end{table}

\subsection{LiDAR Point Cloud Comparison}
\label{sec:appendix-pointcloud}
\begin{figure}[htbp]
\centering
\includegraphics[width=0.62\columnwidth]{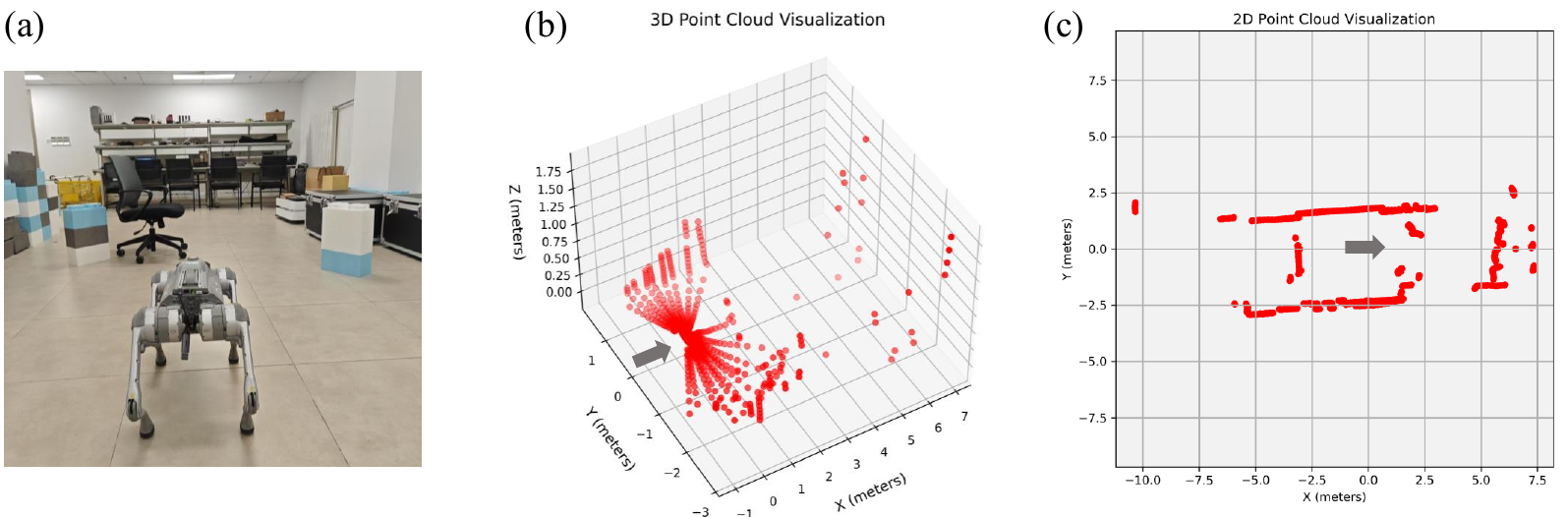}
\caption{Comparison of LiDAR point clouds. The robot's position and orientation are indicated by the gray arrow. (a): The robot's actual environment. (b): Sparse point cloud from the Unitree Go2 robot's onboard LiDAR, consisting of approximately 1,000 3D points per frame. (c): Denser point cloud from the RPLIDAR A2, consisting of approximately 2,000 2D points per frame.}
\label{fig:pointcloud}
\end{figure}

We compare the native Unitree LiDAR L1 and the external RPLIDAR A2 used in our two deployment schemes. The L1 provides a sparse, low-cost 3D point cloud that is directly supported by the onboard controller, enabling plug-and-play deployment. The A2 provides a denser 2D scan that improves near-obstacle geometry for higher-speed navigation. Fig.~\ref{fig:pointcloud} illustrates the qualitative difference between the two sensing streams.

\FloatBarrier

\subsection{Domain Randomization}
\label{sec:appendix-domain-random}
\begin{table}[H]
\centering
\footnotesize
\setlength{\tabcolsep}{3pt}
\renewcommand{\arraystretch}{1.0}
\caption{Domain-randomization parameters.}
\begin{tabularx}{\columnwidth}{@{} L L @{} }
\toprule
\textbf{Term} & \textbf{Value} \\
\midrule
Rays delay                          & $\mathcal{U}(40, 80)\,\mathrm{ms}$ \\
Gravity noise                       & $\mathcal{U}(-0.05, 0.05)$ \\
Linear velocity noise               & $\mathcal{U}(-0.1, 0.1)\,\mathrm{m/s}$ \\
Angular velocity noise              & $\mathcal{U}(-0.1, 0.1)\,\mathrm{rad/s}$ \\
Friction coefficient factor         & $\mathcal{U}(-0.2, 1.25)$ \\
Mass perturbation                   & $\mathcal{U}(-1.5, 1.5)\,\mathrm{kg}$ \\
\bottomrule
\end{tabularx}
\label{table:domain-random}
\end{table}

\FloatBarrier

\subsection{Hyperparameters}
\label{sec:appendix-hyperparams}
Table~\ref{table:train_hyperparams} lists the hyperparameters of ACSI, the LSE-CBF layer, the loss terms, and PPO.
\begin{table}[H]
\centering
\footnotesize
\setlength{\tabcolsep}{3pt}
\renewcommand{\arraystretch}{1.0}
\caption{Hyperparameters for training.}
\label{table:train_hyperparams}
\resizebox{\columnwidth}{!}{%
\begin{tabular}{@{} l l l l @{}}
\toprule
\textbf{Term} & \textbf{Value} & \textbf{Term} & \textbf{Value} \\
\midrule
Buffer length $N_{buf}$   & 100 steps ($\approx$2\,s) & Discount $\gamma$        & 0.998 \\
Min.\ history for replay $\Delta_{min}$ & 20 steps (0.4\,s) & GAE $\lambda$      & 0.95 \\
Rollback $\Delta$ (before clipping) & $\mathcal{U}\{100,\dots,149\}$ steps & Entropy coefficient & 0.003 \\
$p_{replay}$              & 0.8                       & Desired KL               & 0.01 \\
$P_{min}$ / $P_{max}$     & 0.1 / 0.5                 & Init.\ noise std         & 1.5 \\
$d_{up}$ / $d_{down}$     & 0.5\,m / 2.0\,m           & Episode duration         & 60\,s \\
$L_{goal}$ range          & $[0, 10]$                 & $k$ (LSE)                & 10 \\
$d_{safe}$                & 0.20\,m (0.15 + 0.05)     & $\epsilon_d$             & 1.0 \\
$\alpha_{min}$            & 1.0                       & $\lambda_{shield}$       & 0.1 \\
$\lambda_{reg}$           & 1.0                       & $\lambda_{\pi}$ / $\lambda_{V}$ & 0.05 / 0.005 \\
$u^{min}$                 & $[-0.5, -0.8, -1.0]$      & $u^{max}$                & $[1.7, 0.8, 1.0]$ \\
\bottomrule
\end{tabular}%
}
\end{table}

\FloatBarrier
%

\bibliographystyle{IEEEtran}
\bibliography{bib/references}

\end{document}